\documentclass[final]{llncs}
\usepackage[T1]{fontenc}
\usepackage[latin1]{inputenc}
\usepackage{color}
\usepackage{amsfonts}
\usepackage{graphicx}
\usepackage{a4} 

\begin{document}
\title{The observational roots of reference \\of the semantic web\thanks{Dagstuhl Seminar 12221: Cognitive Approaches for the Semantic Web, 28.05.12 -- 01.06.12, Dedre Gentner, Pascal Hitzler, Kai-Uwe Kühnberger, Frank van Harmelen}}
\author{Simon Scheider\inst{1} , Krzysztof Janowicz\inst{1} and Benjamin Adams\inst{2}}
\institute{
Geography, University of California, Santa Barbara 
\email{simonscheider@web.de}
\email{jano@geog.ucsb.edu}
\and Computer Science, University of California, Santa Barbara
\email{badams@cs.ucsb.edu}
}
\maketitle

\begin{abstract}

Shared reference is an essential aspect of meaning. It is also
indispensable for the semantic web, since it enables to weave the global
graph, i.e., it allows different users to contribute to an identical referent. For example, an essential kind of referent is a geographic place, to which users may contribute observations. We argue for a human-centric, operational approach towards reference, based on respective human competences. These competences encompass perceptual, cognitive as well as technical ones, and together they allow humans to inter-subjectively refer to a phenomenon in their environment. The technology stack of the
semantic web should be extended by such operations. This would allow establishing new kinds of observation-based reference systems that help constrain and integrate the semantic web bottom-up.  

\end{abstract}

\setcounter{footnote}{0}
\section{Meaning and Reference}

In the tradition of sciences such as philosophy and linguistics, semantics is the study of meaning, whereas syntax is the study of symbol manipulation. The classical distinction reflects that the realm of meaning is different and should not be confounded with the realm of syntax. 

The things symbols stand for are called \textit{referents}. The notion of meaning involves such essential problems as how humans come to \textit{share} language referents, and how we can \textit{refer} to them in an inter-subjective way. This problem is what we call the \textit{problem of reference}. In many respects, it can be considered a practical manifestation of the \textit{symbol grounding problem} \cite{Harnad.1990}, the problem of how to get out of the realm of syntax into any determinable contact with referents in the world. The problem of reference has been an ongoing issue in the semantic web \cite{HayesH08}, and symbol grounding recently was recognized as an untackled problem  \cite{cregan:symbol}. Like symbol grounding, the problem of reference is not solved by formal semantics of a Tarskian flavour, because formal theories can neither distinguish a particular domain nor a particular interpretation \cite{Scheider.2009b}. Furthermore, natural language descriptions cannot account for reference, too, because they are context dependent, in particular since they rely on polysemy and synonymy \cite{Scheider.2009b}.

The semantic web was proposed to make the meaning of information on the web explicit. Reference is an essential part of meaning that is also indispensable for the semantic web, especially for linked data. In particular, it is shared reference which allows to establish identity, 
and, thus, allows different users to contribute information to the same information item, e.g., different tags to the same place in Linked Geo Data. This so called \textit{network effect} is critical to the semantic web \cite{booth2006}. However, often, semantic web research is driven by syntax and technology stacks in the first place. 

We argue for enlarging the breadth of semantic web research by focusing on the problem of reference. This requires us to learn from existing grounded symbol systems and to notice how humans cognize \cite{RaubalAdams10swj} and disambiguate referents in language.
\section{An operational view on the roots of reference}
We argue for an operational view on the problem of reference and take up on the ongoing debate in the semantic web.

\textit{Meaning and referencing cannot be reduced to symbols.}
Some authors have tried to account for reference in terms of URIs \cite{Berners-LeeFischetti:1999}, formal semantics \cite{parsia:meaning,presutti:identity}, and appropriate descriptions \cite{booth2006}.
Such approaches rely on the assumption that while meaning is external to the web, it can be regarded as internal to the semantic web; see also \cite{parsia:meaning,Berners-LeeFischetti:1999}. 
However, it is important to note that while the language of the semantic web is machine readable, meanings are not.   
Hayes points out that it is not possible to solve the problem of reference within language constructs \cite{HayesH08}. 
The tendency seems therefore to surrender and start praising the benefits of ambiguity in language instead \cite{HayesH08,parsia:meaning}. 
The existence of ontological disagreement \cite{parsia:meaning}, however, does not mean that we can dispense with intended meaning, as Ginsberg argued \cite{ginsberg:ontological}. Intended meanings provide reference, and shared reference is critically indispensable in order to prevent data isolation. 

\textit{Referencing is something human observers do.}
We argue that there is no need to surrender but a need to shift perspective on the problem. Hayes and Halpin hint at the right direction when they state that ``reference is part of a communicative act between cognitive agents'' that involves conventions \cite{HayesH08}. We argue that referencing (and meaning as well) is not an object or some abstract mapping, but something that humans do \cite{Kuhn.2009}, namely a kind of \textit{speech act} that draws human attention to some reproducible phenomenon \cite{Scheider.2011b}. However, one should be aware that understanding this act draws on almost the full range of human cognitive capabilities\footnote{See the discussion in \cite{Scheider.2011b}.}, including Gestalt perception, joint attention \cite{Tomasello.1999}, conventionally established predications\footnote{See \cite{Kamlah.1996}. Quine's \textit{observation sentences} \cite{Quine.1974} reflect a similar idea.} and imagination.  

\textit{Technical language reference is based on human operations.}
The problem of where to begin the construction of a concise technical language \cite{Kamlah.1996}, i.e., where to ground terms \cite{Scheider.2011b}, is decisive for its meaningfulness, and is much too often disregarded in information ontology. \cite{Kamlah.1996} have argued that in order to prevent talking at cross purposes, it is necessary to begin with a language level whose interpretation is internalized, embodied in practice, and therefore shared among speakers, and on which a more concise technical language can be constructed. Paul Lorenzen argued that in order to understand the method of thought, we need to comprehend the extra-linguistic actions that are needed to build a language \cite{Kamlah.1996}.


Does this mean that the current semantic web technology stack is useless? No, since ontologies are useful to constrain the interpretation and to infer implicit knowledge. However, they constrain interpretation only top-down, while shared references are based on human capabilities that constrain interpretation bottom-up \cite{Kuhn.2009}. We argue that the \textit{technology stack of the semantic web needs to draw on these cognitive capabilities} in order to allow bottom-up semantic constraints \cite[c.f.]{Janowicz.2012}. 

Based on this argumentation, we have shown in previous work how to reconstruct \textit{geospatial concepts} based on perceptual operations and constructions instead of describing them top-down \cite{Scheider.2011b,AdamsJanowicz2011}. For example, geographic places \cite{Scheider.2010d}, water bodies and environmental media \cite{Scheider.2009}, environmental qualities \cite{Scheider.2011}, road networks and junctions can be reconstructed based on observing \textit{action potentials} (affordances) \cite{Scheider.2010a}. 


\paragraph{Reference systems.}
Solutions to the problem of reference should transgress syntax as well as technology. They cannot solely rely on computers but must also rely on human referential competences. This requirement is met by \textit{reference systems} \cite{Kuhn.2003}. Reference systems are different from ontologies in that they constrain meaning bottom-up \cite{Kuhn.2009}. Most importantly, they are not "yet another chimaera" invented by ontology engineers, but already exist in various successful variants. A prominent example are \textit{spatial (geodetic) reference system}, which provide a language about points on a mathematical ellipsoid anchored relative to the earth surface by reference observations of places and directions. It is responsible for the success of geographic information systems (GIS), navigation systems, and serves as an integration layer for the semantic web \cite{Janowicz.2010}. Another example are \textit{measurement scales}, whose units of measurement are likewise established by conventional observations. \textit{Address systems} are (geo-)reference systems for certain kinds of observable places, namely buildings. \textit{Calendars} are temporal reference systems anchored in the observation of day and night, as well as celestial bodies. 

\paragraph{New kinds of reference systems as semantic integration layer.} A reference system can be seen as a theory in which abstract or vague concepts are reconstructed based on conventionally established and, therefore, reproducible observation procedures \cite{Scheider.2011b}. This makes it useful as a semantic plane for linking heterogeneous conceptualizations that still divide the semantic web. A comparable idea is Peter Gärdenfors' notion of a conceptual space \cite{Gardenfors.2004}. This is a geometric, topological space whose dimensions are anchored in human cognition, and in which concepts are expressed as convex regions. For example, local conceptualizations of a term like ``high mountain'' can be compared in terms of different convex regions in this space \cite{AdamsJanowicz2011}. New kinds of reference systems would allow us to negotiate or translate between heterogeneous views on the world \cite{Janowicz.2010}. For example, places are essential concepts for spatial referencing of user generated content on the social web, such as points of interest (POI), geo-tagged pictures, blog posts or tweets. But they are highly vague both regarding their boundaries as well as regarding category labels \cite{Scheider.2011b}. A \textit{place reference system} would allow us to disambiguate place concepts beyond spatial footprints, taking into account the different types of afforded actions that constitutes them as well as their spatio-temporal distribution \cite{Scheider.2010d}. 

\section{Conclusion}
We argued that the technology stack of the semantic web should additionally draw on human operations that account for reference. This would enable us to establish and reconstruct reference systems, beyond the existing ones for measurements, locations, times and buildings. What we need are more and other kinds of reference systems in order to account for the various semantic domains involved in the semantic web. These may allow us, for example, to construct concepts as abstractions in terms of logical reifications or convex regions in conceptual spaces. They assure inter-subjectivity of reference based on joint human observation procedures and convention \cite{Scheider.2011b}, but require us to broaden the focus of semantic research from top-down, language-centric to bottom-up, human-centric and operational approaches.  
\bibliographystyle{splncs}

\end{document}